\documentclass{bmvc2k}

\usepackage{amsfonts,amssymb} 
\title{Partition-based Nonrigid Registration \\for  3D Face Model}

\addauthor{Yuping Ye $^1\,\,^3$ }{yp.ye@siat.ac.cn}{1}
\addauthor{Zhan Song $^1\,\,^2$}{zhan.song@siat.ac.cn}{2}
\addauthor{Juan Zhao $^1\,\,^2$}{juan.zhao@siat.ac.cn}{2}

\addinstitution{$^1\,$
Shenzhen Institutes of Advanced Technology, Chinese Academy of Sciences.\,\,Shenzhen, China.
\\
$^2\,$
The Chinese University of Hong Kong.\,\,Hong Kong,China.
\\
$^3\,$
University of Chinese Academy of Sciences.\,\,Beijing,China.
}

\runninghead{Ye et al}{Partition-based Nonrigid Registration for  3D Face Model}

\def\etal{\emph{et al}\bmvaOneDot}

\begin{document}

\maketitle

\begin{abstract}
This paper presents a partition-based surface registration for 3D morphable model(3DMM). In the 3DMM, it often requires to warp a handcrafted template model into different captured models. The proposed method first utilizes the landmarks to partition the template model then scale each part and finally smooth the boundaries. This method is especially effective when the disparity between the template model and the target model is huge. The experiment result shows the method perform well than the traditional warp method and robust to the local minima.
\end{abstract}

%-------------------------------------------------------------------------
\section{Introduction}
\label{sec:intro}

Registration is a fundamental task in computer vision and computer graphics. Works related to the registration can be classified into two classes: the rigid registration\cite{besl1992method,sharp2002icp,yang2016go} and the non-rigid \cite{allen2003space,amberg2007optimal,ma2016non}. The research on rigid registration aim at seeking an appropriate transformation which best aligned one point cloud/mesh to the other. Compared to the rigid registration, the nonrigid work in quite a different way. The process of the nonrigid registration can be view as warping the template model onto the target. Every point from the template aligns to the target often in different transformation parameters. Nonrigid registration play a pivotal role in  3D reconstruction and animation. While 3D morphable model (3DMM) as the most representative work in facial animation was first proposed by 
Blanz and Vetter\cite{blanz1999morphable}. 3DMM algorithm reconstructs the face geometry or texture from a single image or sparse point clouds based on the 3D statistical models generated from the captured 3d face database.   In order to get the statistical information from the captured 3D face database, each captured mesh must be re-parameterized into a consistent form where the number of vertices, the facets and the anatomical meaning of each vertex must be made consistent across all meshes\cite{booth2018large}.  Nonrigid registration which is one of the common re-parameterize methods in 3DMM warps one template model into all the captured face models.  The performance of the nonrigid registration method directly affects the quality of the model generated by 3DMM algorithm.

To achieve the accuracy result by 3DMM, we need to secure the re-parameterize method to warp the mesh corresponding anatomically accurate.   For acquiring canonical statistical information, the template model is often hand-crafted.  The challenge of the re-parameterize method is that we have to warp the hand-crafted model into all captured mesh varied from age, gender and ethnicity.  The traditional re-parameterize method often failed when the disparity between the template model and the target is large. In this paper, we proposed a partition-based re-parameterize method performs well for this hard nut. 

There have been many works that contribute to re-parameterizing the captured models into the consistent form. Those re-parameterization methods can be divided into two classes: the projection-based method \cite{blanz1999morphable,cosker2011facs,patel20093d} and the registration-based \cite{amberg2007optimal, allen2003space, booth20163d}. 

The projection-based re-parameterize methods demand the projection relationship between the texture image and the mesh beforehand. These methods often cannot reach the anatomical consistency. The re-parametrize method described in the pioneering work\cite{blanz1999morphable} projected each facial model into a virtual cylindrical texture image then sampled the cylindrical texture image uniformly. This method failed to re-parameterize the different size organs into the same number of triangles.  In \cite{patel20093d}, the authors manually annotated the cylindrical face projections with a set of sparse annotations, then employed a thin plate spline to warping the texture into a standard reference one. 

The registration-based methods warp the template model into the target by minimizing the proposed objective function.  These methods often have to overcome problems such as converging slowly and falling into the local minimum easily. In \cite{allen2003space}, the authors utilized the objective function with distance, smooth and markers regulations to fitting high-resolution template model to detailed human body range scans. Amberg \etal \cite{amberg2007optimal} extended the work of \cite{allen2003space} proposed the optical step nonrigid ICP framework for surface registration. In \cite{paysan20093d}, the optimal step nonrigid ICP algorithm was introduced to warping the template into scans of 200 subjects. In \cite{booth2018large}, the authors also adopted the algorithm in \cite{amberg2007optimal} algorithm to construct the 3D morphable model from a large scale of facial identities. The method we proposed also belongs to this class.

The organization of this paper is as follow. The pitfalls of the traditional registration-based method are analysis in Section $2$. The proposed partition-based nonrigid registration is introduced in Section $3$. Experimental results are provided and discussed in Section $4$. Conclusion and future works are offered in Section $5$.

\section{Pitfalls of Traditional Registration-based Method}
As the above mentioned, the nonrigid registration-based re-parametrize method, in essence, is an energy minimization problem. An optimization problem can be view as a two-step process: first, established the energy function; second, employed the suitable optimization algorithm to find the solution. The hand-crafted template model $M_{template}=	\{V_{template} \in \mathbb{R}^{3n},$  \\ $E_{template}	\in\left\{V_{template},V_{template}\right\}\}$ where $V_{template} $ denote the $n$ vertices and $E_{template} $ denote the $m$ edges
 is often a manifold triangular mesh.While the target model $ M_{target}$ can be given in any representation such as point cloud, triangle mesh which merely needs to meet the requirement of allowing being found the closest position to the template vertices. The warping process transforms each template point $\left\{V_i\in \mathbb{R}^{3}\right\}$ to matching with the target model via the affine transformation $\left\{T_i\in \mathbb{R}^{3\times4}\right\}$. The energy function adopted in \cite{allen2003space,amberg2007optimal,booth2018large} is composed of three parts: distance regulation, stiffness regulation, and landmarks regulation.
 \begin{equation}
\begin{aligned}
 min\,\,\,E(T)=\alpha E_d(T)+\beta E_s(T)+\gamma E_l(T)
 \end{aligned}
\end{equation}
where:
\\
\\
 Distance \,\,\,\,\,\,Term\,\,$E_d(T)$:This aim of this term is to make the template point close to the target as soon as possible.\,\,\,$E_d=\sum_{i=1}^n{\left\|T_i^t \left[ \begin{matrix}
   V_i\\
   1 \\
  \end{matrix}\right]-\left[ \begin{matrix}
   V_{closest\_i}\\
   1 \\
  \end{matrix}\right]\right\|^2}$ \,\,\,where $ V_{closest\_i}$ indicates the closest position of the target to the template point $V_i$.
 \\
 \\
 Stiffness \,\,\,\,\,\,Term\,\,$E_s(T)$: Only the distance regulation may result in an irrational mesh. This term restrains the neighbor vertices transformation from shifting too dramatic and unreasonable.\,\,\,\,$E_s=\sum_{\left\{i,j|\{V_i,V_j\}\in E_{template}\right\}}^m{\|T_i-T_j\|^2}$ This term penalizes all the transformation difference of all the edge's vertices for a smooth mesh.
 \\
 \\
 Landmarks Term\,\,$E_l(T)$: Similar to the Distance term, this term avoids the solution stuck in the local minima.\,\,\,$E_l=\sum_{i=1}^k{ \left\|T_{\kappa(i)}^t \left[ \begin{matrix}
   V_{\kappa(i)}\\
   1 \\
  \end{matrix}\right]-\left[ \begin{matrix}
   LM_i\\
   1 \\
  \end{matrix}\right]\right\|^2}$\,\,where The $LM_i$ indicates the position the $\kappa(i)\,th$ vertice should be placed.
\begin{figure}[htb]
\begin{minipage}[b]{1.0\linewidth}
  \centering
  \centerline{\includegraphics[width=9cm]{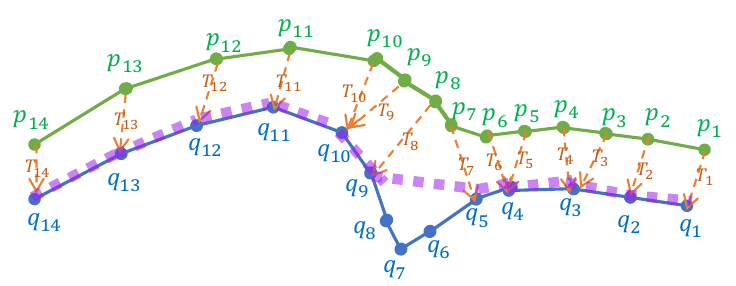}}
\end{minipage}
\caption{Example of failure of the traditional registration-based method.}
\label{fig:res}
\end{figure} 

When the disparity between the template model and the target is huge, the algorithm has a long way to go and easy to trap in local minima.Although the landmarks regulation can avoid falling the local minima to some extent, yet it often doesnot work in many cases.We can explain it with an example. As Figure 1 shows, the green line indicates the template and the blue line for the target. The orange arrows direct the closest vertices of the target from the template. Because of the relationship of these vertices and transformations shown as follow.
\begin{equation}
\begin{aligned}
&\alpha\sum_{i=4}^{10}{\left\|\hat{T}_i^t \left[ \begin{matrix}
   p_i\\
   1 \\
  \end{matrix}\right]-\left[ \begin{matrix}
   q_i\\
   1 \\
  \end{matrix}\right]\right\|^2}+\beta\sum_{i=4}^{10}{\left\|\hat{T}_i-\hat{T}_{i-1}\right\|^2}+\gamma\left\|\hat{T}_7^t \left[ \begin{matrix}
   p_7\\
   1 \\
  \end{matrix}\right]-\left[ \begin{matrix}
   q_7\\
   1 \\
  \end{matrix}\right]\right\|^2
  \ge 
  \\&
  \alpha\sum_{i,j=4,1}^{10,7}{\left\|\tilde{T}_i^t \left[ \begin{matrix}
   p_i\\
   1 \\
  \end{matrix}\right]-\left[ \begin{matrix}
   q_{\Phi(j)}\\
   1 \\
  \end{matrix}\right]\right\|^2}+\beta\sum_{i=4}^{10}{\left\|\tilde{T}_i-\tilde{T}_{i-1}\right\|^2}+\gamma\left\|\tilde{T}_7^t \left[ \begin{matrix}
   p_7\\
   1 \\
  \end{matrix}\right]-\left[ \begin{matrix}
   q_7\\
   1 \\
  \end{matrix}\right]\right\|^2
    \\&where:\,\,\,\,\Phi=\left\{3,4,4,5,9,10,10\right\}
\end{aligned}
\end{equation}
The solution of the nonrigid registration is trap in the local minima $\tilde{T}$, and the purple line shown in Figure 1 is the final warping result. We will give a real example of the failure of the traditional registration-based method in Sec. 4.

\section{Partition-based Nonrigid Registration}

To solve the above-presented problem, we put forward two strategies to shorten the warping methods' minimization journey and find the global minima.  One is to partition and preprocess the template model, and the other is to revise the energy function. 

A good 3DMM algorithm must warp the template model into a consistent form according to the scanned meshes varied from gender, age, and ethnicity. Take the nose as an instance. Because of the climate and geography difference,  European often have sharp and higher nose while Asian's nose is often lower and their nasal tip is flat. To reach the anatomical meaning consistent,  we have to warp the template nose into these distinctive noses. In the 3DMM  algorithm, all the facial organs in the template model need to match the target accurately. Based on the analysis, we first partition the template model into several parts via the landmarks; then scale each part separately, and finally smooth the boundary between the parts.

\textbf{Partition the template}\,\,\,
To partition the template model, we need to have several landmarks on the template model.  These landmarks $ \left\{V_{\kappa(i)}\right\}$ are often given manually on the hand-crafted template model. According to the distribution of the facial organs, we can line some landmarks and give the partition strategy. Figure 2 depicted two strategies for template partition. Figure 2 c). shows the template model is divided into two parts by the 11 handpicked landmarks, while Figure 2 d). shows the template model is separated into eight parts by the 68 Dlib landmarks. 

\begin{figure}[htb]
\begin{minipage}[b]{1.0\linewidth}
  \centering
  \centerline{\includegraphics[width=12cm]{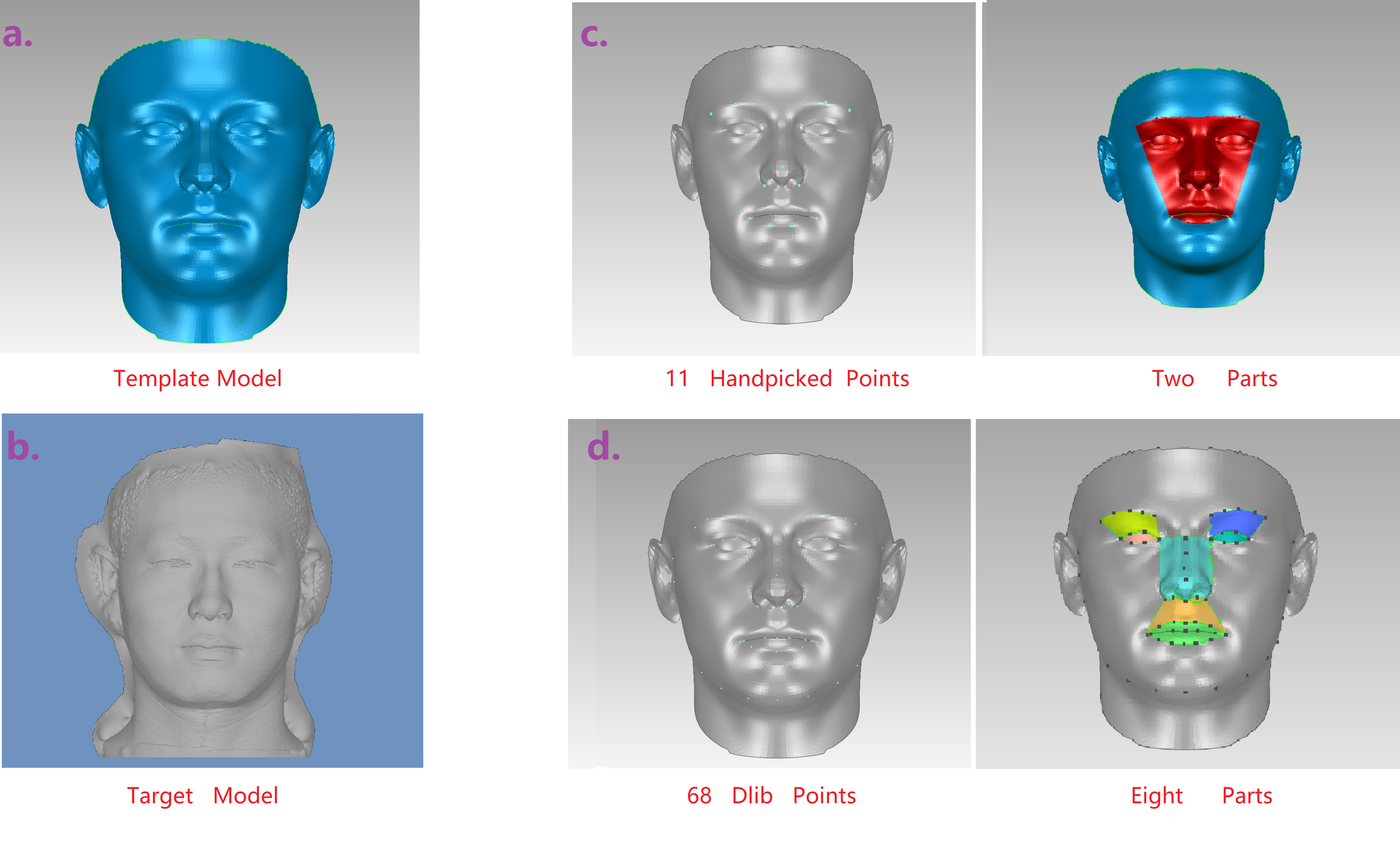}}
\end{minipage}
\caption{\,\,a).\,\,The template model modified from \cite{paysan20093d}. \,\,b).\,\,One target model captured by \cite{song2013accurate}.\,\,c).\,\,Template partition by the handpicked points.\,\,d).\,\, Template partition by the Dlib \cite{dlib09} points. }
\label{fig:res}
\end{figure} 

\textbf{Scale each part}\,\,\,
In this step, we will scale each part of the template model to approximate the target via an affine matrix\,\,$X_i\in \mathbb{R}^{4\times3}$. The landmarks on the target model $\left\{LM_i\right\}$  can be selected in two ways: one is to handpick the landmarks by the aid of the 3D visualization software, the other is to detect the texture automatically via the facial landmarks algorithm  then use the 3D-2D relationship to find the matched landmarks. To obtain the affine matrix of each part, we need to have several landmark pairs. We implement the landmark pairs on the $ ith$ part of the template model $ A_i=\left\{\left[ \begin{matrix}V_{\kappa(i)}\in P_{template}^i \\1\end{matrix}\right]\right\}$ and the target $B_i=\left\{LM_i\in P_{target}^i \right\}$  to form the linear matrix equation:\,\,\,\,$A_i^t X_i=B_i^t$ \,\,where:\,\, $P_{template}^i,\,\,
P_{target}^i$ respectively indicate the $ith$ part of the template model and the target. Then the SVD method is adopted to solve these equations.  Finally, we apply the affine matrix to scale each part and get the scaled model:
\begin{equation}
\begin{aligned}
&M_{template}^{scale}=\left\{X_{\zeta(i)}^t
                      \left[ \begin{matrix}
                                     V_i\\1
                              \end{matrix}
                      \right]
               \right\} 
               \\
               &\,\,\,where\,\,\zeta(i)\,\,\,represents \,the\, part\, where\, ith\, vertice\, on. 
\end{aligned}
\end{equation}

The vertices near the boundary among the parts may shift rather dramatical. To assure the energy function especially the stiffness item work efficient, we have to smooth the boundaries. Each boundary line $l_{i,j}$ corresponds to two part $P_{template}^i,\,\,P_{template}^j$\,\,\,. If the distance\,\, $d_{v_i \mapsto l_{\zeta(i),\chi(i)}}$ \,\,from the vertice \,$V_i$\, to its' nearest boundary line \,$l_{\zeta(i),\chi(i)}$\, is less than $\Delta$,  the vertice coordinate will be adjusted to : $$V_i^{smooth}=(\frac{1}{2}+\frac{d_{v_i \mapsto l_{\zeta(i),\chi(i)}}}{2\Delta})
X_{\chi(i)}^t\left[ \begin{matrix}
                                     V_i\\1
                              \end{matrix}
                      \right]+(\frac{1}{2}-\frac{d_{v_i \mapsto l_{\zeta(i),\chi(i)}}}{2\Delta})X_{\zeta(i)}^t\left[ \begin{matrix}
                                     V_i\\1
                              \end{matrix}
                      \right]$$ 
   where $\chi(i)$ indicates the other corresponded part of the $ith$ vertice's nearest boundary line.
    Till now, we obtain the final template model:
\begin{equation}
\begin{aligned}
&M_{template}^{final}=
\left\{
V_i^{final}=
\begin{cases}X_{\zeta(i)}^t
\left[ \begin{matrix}
                                     V_i\\1
                              \end{matrix}
                      \right]
                      & \text{ $d_{v_i \mapsto l_{\zeta(i),\chi(i)}}\ge\Delta$ }\\
                      \\
V_i^{smooth}& \text{$d_{v_i \mapsto l_{\zeta(i),\chi(i)}}<\Delta$}
\end{cases}
\,\,\,\,\,\,
\right\}
\end{aligned}
\end{equation}

The above three steps can significantly shorten the route to the solution especially when the disparity is large. Besides the distance regulation, stiffness regulation and landmarks regulation, we introduce normal regulation to avoid the risk of falling into local minima. 
\begin{equation}
\begin{aligned}
 min\,\,\,E(T)=\alpha E_d(T)+\beta E_s(T)+\gamma E_l(T)+\eta E_n(T)
\end{aligned}
\end{equation}
 Normal \,\,\,\,\,\,Term\,\,$E_n(T)$:\,\,This term similar to the Distance term aim to regulate the normal change dramatically.\,\,\,$E_n=\sum_{i=1}^n{\left\|T_i^t \left[ \begin{matrix}
   N_i\\
   0 \\
  \end{matrix}\right]-\left[ \begin{matrix}
   N_{closest\_i}\\
   0 \\
  \end{matrix}\right]\right\|^2}$ \,\,\,where $ N_{closest\_i}$ indicates the closest point's normal of the target to the template point $V_i$.
\section{Experimental Results}

At the start of the experiment, we have to gather some 3D models to evaluate our algorithm.  We remove the tongue part of the model from \cite{paysan20093d} as the template model. In the practical application \cite{yu2017robust,mora2012gaze},  the target model is often captured by the commercial off-the-shelf device such as Kinect and Realsense. In order to better evaluate our algorithm, we adopt the accurate and dense 3D models generated by the structured light system \cite{song2013accurate} as the target models. When the target model is accurate, and of details, we can evaluate the warp method directly by the visual effect.

\begin{figure}[ht]
\begin{minipage}[b]{1.0\linewidth}
  \centering
  \centerline{\includegraphics[width=7cm]{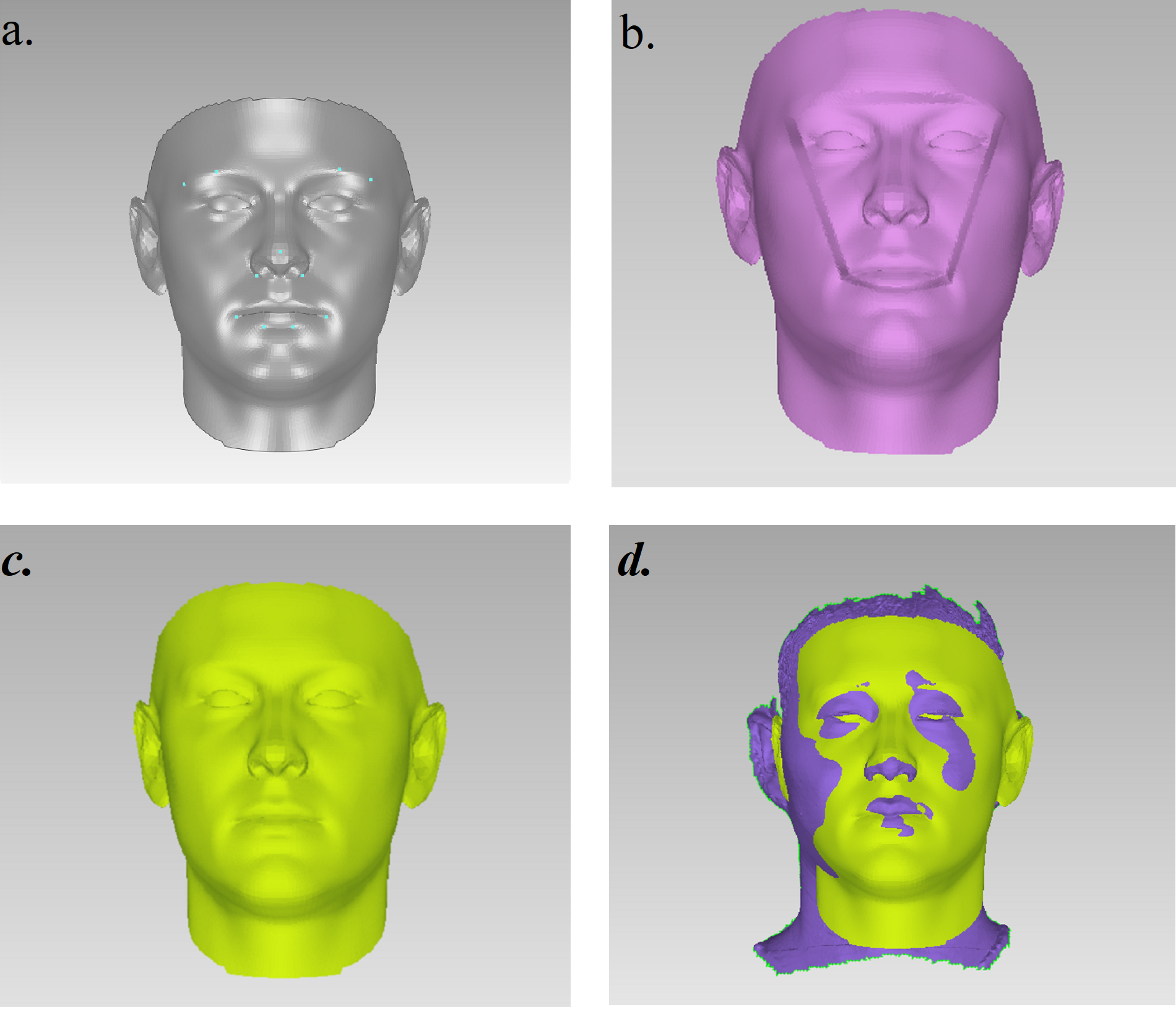}}
\end{minipage}
\caption{Steps of the partition-based algorithm }
\label{fig:res}
\end{figure} 
As Figure 3 a). to  c). shows, we first select 11 landmark points of the template and the target and partition the models into two parts; Then, each part of the template model is scaled respectively via the Equation 3; Finally, we smooth the boundaries across the parts.  Figure 3 d). shows the deviation between the scaled template model and the target.  Finally, we adopt the optimization method proposed in \cite{amberg2007optimal} to minimize our method energy function presented in Equation 5.
\begin{figure}[ht]
\begin{minipage}[b]{1.0\linewidth}
  \centering
  \centerline{\includegraphics[width=9cm]{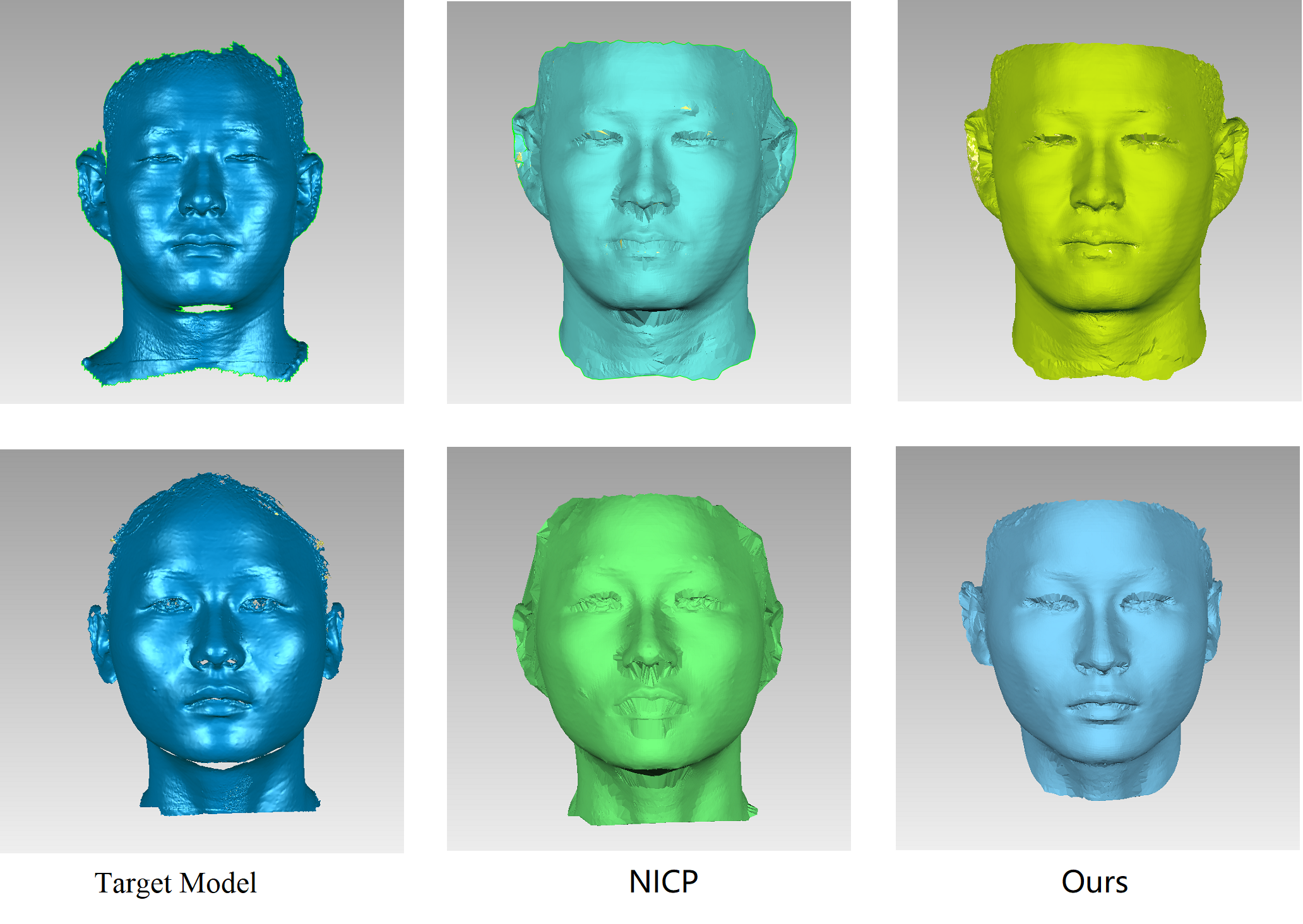}}
\end{minipage}
\caption{Comparision of registration quality between NICP and ours.}
\label{fig:res}
\end{figure} 

To illustrate the proposed method is robust to the local minimas, we conduct a comparative evaluation between the method in \cite{amberg2007optimal} (NICP)  and ours'. As Figure 4 shows, we find the NICP method failed to warp the nose and the mouse correct.  The NICP method trap in the local minima which have been discussed above, while our method performs well in this case. In Figure 5, we adopt the partition strategy shown in Figure 2  to warp the template model into the several captured models.

\begin{figure}[htbp]
\begin{minipage}[b]{1.0\linewidth}
  \centering
  \centerline{\includegraphics[width=12.5cm]{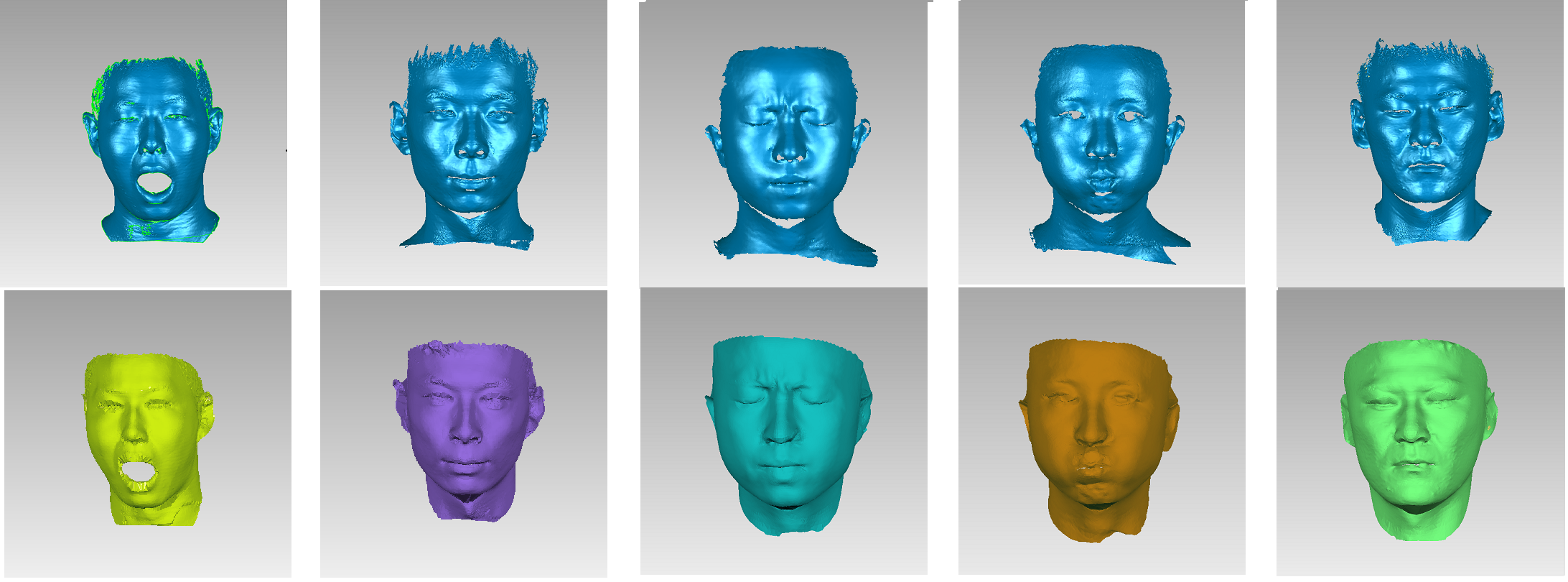}}
\end{minipage}
\caption{Registered 3D models of different facial expression by the 68 Dlib Partition strategy.}
\label{fig:res}
\end{figure}

\section{Conclusion}

In this paper, a partition-based nonrigid registration method is introduced for 3D morphable model reconstruction. We firstly analyze the pitfalls that easy to fall into the local minima of the traditional nonrigid registration method. The proposed method employs the landmarks to partition the template model then scale each part by its' affine matrix and finally smooth the boundaries across the parts. The proposed method can significantly shorten the route to the solution and it robust to the local minima. Finally, we conduct several qualitative experiments to evaluate the proposed method. Future work can address how to implement the proposed method for better 3D morphable model reconstruction.

\bibliography{egbib}
\end{document}